\documentclass[letterpaper]{article} % DO NOT CHANGE THIS
\usepackage{aaai2027}  % DO NOT CHANGE THIS
\usepackage[hyphens]{url}  % DO NOT CHANGE THIS
\usepackage{graphicx} % DO NOT CHANGE THIS
\usepackage{natbib}  % DO NOT CHANGE THIS AND DO NOT ADD ANY OPTIONS TO IT
\usepackage{caption} % DO NOT CHANGE THIS AND DO NOT ADD ANY OPTIONS TO IT
\usepackage{multirow} 
\usepackage{amsmath}
\usepackage{amssymb}
\DeclareMathOperator*{\argmax}{arg\,max}

\usepackage{booktabs}

\title{Rethinking Video Token Compression with a Global Codebook: Learning Once, Compressing Everywhere}
\author{
Jiayang He\equalcontrib$^1$,
Tianling Xu\equalcontrib$^1$,
Diancheng Kang$^1$,
Huaide Jiang$^1$,
Junyan Bai$^1$,
Shaoming Zheng$^2$\corresponding,
Xuan Song$^3$\corresponding
}

\affiliations{
$^1$Southern University of Science and Technology\\
$^2$Imperial College London\\
$^3$Jilin University
}
\begin{document}

\maketitle

\begin{abstract}
Video large language models (Video-LLMs) represent videos as dense sequences of visual tokens, whose length grows with the temporal and spatial extent of the input. These tokens often contain substantial redundancy arising from repeated visual patterns, leading to unnecessary computation in the subsequent language-model processing. Existing token compression methods, including pruning and merging, perform compression online during inference, repeatedly incurring additional computation for each input video and often relying on model-specific designs that limit their generality, we instead rethink this paradigm by shifting the costly compression process offline. We propose \textbf{ONCE}, a plug-in video token compression framework that introduces an offline-to-online paradigm: a frequency-aware global codebook is learned once in the visual feature space and reused for lightweight online compression through codebook lookup and aggregation, reducing repeated per-video computation and the need for model-specific compression designs. Extensive experiments across multiple video understanding benchmarks and against diverse compression baselines demonstrate that our approach achieves a strong accuracy-efficiency trade-off, maintaining competitive performance while achieving the lowest inference latency among compared methods.

\end{abstract}

% Uncomment the following to link to your code, datasets, an extended version or similar.
% You must keep this block between (not within) the abstract and the main body of the paper.
% Make sure that you do not de-anonymize yourself with these links.
% \begin{links}
%     \link{Code}{https://aaai.org/example/code}
%     \link{Datasets}{https://aaai.org/example/datasets}
%     \link{Extended version}{https://aaai.org/example/extended-version}
% \end{links}

\section{Introduction}

Video understanding with large multimodal models requires processing thousands of visual tokens extracted from video inputs. As video duration increases, the resulting sequences place growing pressure on the context window and prefill computation of the language model~\cite{video-llava,llava-onevision,visionzip}. Meanwhile, neighboring frames and patches often contain repeated visual information~\cite{dycoke,framefusion}, making efficient token compression an important problem.
% ----------------------------------
Existing methods mainly compress visual tokens through selection~\cite{fastv,visionzip,llava-prumerge,forestprune}, aggregation~\cite{tome,dycoke,framefusion,li2026token}, or compact representation learning~\cite{llama-vid,vqtoken}. Despite their different designs, most follow an inference-time workflow: dense visual features are first generated and then compressed independently for each input video.
% ----------------------------------
Despite their differences, existing methods perform compression only after dense visual tokens have already been generated. As a result, a substantial portion of the visual encoding cost has already been incurred before compression takes place, making token compression itself an additional inference-stage computation rather than an intrinsic part of visual representation learning.
% ----------------------------------
Moreover, compression decisions are computed independently for every input video. Although videos often share recurring objects, scenes, and visual structures, existing approaches repeatedly estimate token importance or similarity from scratch~\cite{fastvid,forestprune,unified-spatialtemporal,li2026token}, preventing compression knowledge from being shared or reused across different videos.
% ----------------------------------
Finally, many existing approaches rely on model-dependent compression mechanisms, such as attention-based importance estimation or architecture-specific similarity metrics~\cite{fastv,llava-prumerge,prunevid}. Such designs often couple the compression process with the underlying Video-LLM architecture, requiring additional adaptation when deployed across different models and limiting their flexibility in practical scenarios.
% ----------------------------------
\begin{figure*}[t]
\centering
\includegraphics[width=\textwidth]{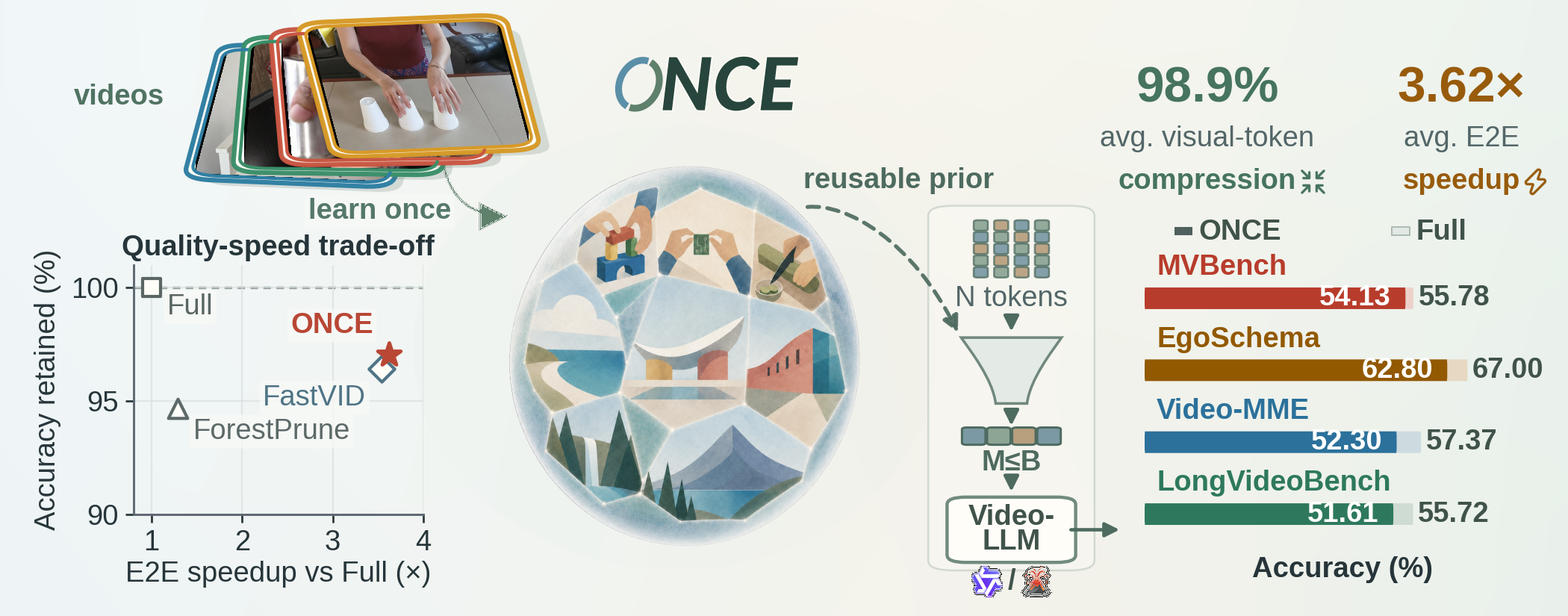}
\caption{ONCE learns a reusable global codebook from an offline video corpus.
For each test video, fixed-codebook assignment and continuous mean pooling
replace $N$ projected visual tokens with at most $B$ pooled tokens before
language-model processing. The plotted LLaVA-OneVision-7B operating point uses
$B=256$ and compares task accuracy with the corresponding Full model.}
\label{fig:once_teaser}
\end{figure*}

% ----------------------------------
In this work, we propose \textbf{ONCE}, an offline-to-online video token compression framework (Figure~\ref{fig:once_teaser}) that learns a reusable grouping prior in a frozen visual feature space. Offline, ONCE fits a balanced global codebook to temporally stratified sketches weighted by source-token count. During inference, it retains the codewords best supported by the current video, reassigns all source tokens to them, and mean-pools their original continuous embeddings. The codebook guides grouping while all Video-LLM parameters remain frozen.

Our main contributions are summarized as follows:

\begin{itemize}

    \item We introduce an novel offline-to-online framework that learns a corpus-level grouping prior and reuses it for unseen videos in the same frozen feature space.

    \item We learn a global codebook from source-count-weighted sketches, with budget-aware selection and source-token mean pooling.

    \item We evaluate ONCE across four video understanding benchmarks and two model backbones against input-adaptive compression baselines. ONCE achieves a favorable accuracy--efficiency trade-off, with its strongest accuracy gains under tight token budgets and consistent reductions in inference cost.

\end{itemize}

\section{Related Work}
\subsection{Video Large Language Models}

Recent advances in multimodal large language models have enabled Video-LLMs to perform complex video understanding tasks by integrating pretrained visual encoders with large language models. Early Video-LLMs, such as Video-ChatGPT and Video-LLaVA, establish a modular architecture that connects pretrained vision encoders with LLMs through lightweight projection modules~\cite{video-chatgpt,video-llava,video-llama,videochat}. Recent works, including LongVILA, LongLLaVA, and LLaVA-OneVision, further scale Video-LLMs toward long-video understanding by extending visual contexts, improving temporal reasoning, and leveraging large-scale video instruction data~\cite{longvila,longllava,llava-onevision,llava-onevision-1.5}. Meanwhile, general-purpose vision-language models such as InternVL and Qwen3-VL provide stronger multimodal representations and flexible visual token processing, serving as powerful backbones for video understanding tasks~\cite{internvl,internvl3,qwen3vl}. However, these advances rely on increasingly dense visual-token sequences, where the growing number of visual tokens becomes a major bottleneck for long-video inference.

\subsection{Video Token Compression}

Existing video token compression methods reduce redundant visual tokens through different compression operations, mainly including token pruning and token aggregation. Token pruning methods remove less informative tokens by estimating token importance or redundancy, reducing the number of visual tokens while preserving essential information~\cite{fastvid,forestprune,hieravid,prunevid,metok}. In contrast, token aggregation methods merge redundant tokens into compact representations by exploiting feature similarity, temporal redundancy, or token correspondence~\cite{tokenfusion,dycoke,framefusion,holitom}.

These compression operations typically rely on different token decision proxies to estimate token importance or redundancy. Existing approaches leverage various signals, including spatio-temporal structures~\cite{forestprune,hieravid}, temporal and visual redundancy~\cite{fastvid,dycoke}, feature similarity~\cite{tokenfusion,framefusion}, language-query relevance~\cite{prunevid}, and information-aware criteria~\cite{infomerge}.

Despite using different compression operations and decision proxies, existing methods generally follow an instance-specific compression paradigm, where compression decisions are determined independently for each input video after dense visual tokens have been generated. As a result, compression patterns need to be repeatedly estimated for each video, limiting the reuse of compression knowledge across different inputs and motivating the exploration of reusable compression priors.

\subsection{Codebook-based Representation Learning}

Codebook-based representation learning learns a finite set of representative prototypes to discretize continuous visual features into compact latent spaces. Early approaches such as VQ-VAE and VQGAN introduce vector quantization to construct discrete visual representations for image reconstruction and generation~\cite{vqvae,vqgan}. Recent works extend this idea to multimodal models by learning discrete visual vocabularies, where continuous visual features are mapped to compact token representations for efficient visual understanding and generation~\cite{vqtoken,beingvl0.5}.

Inspired by these discrete representation learning approaches, we explore a different perspective of codebook utilization for video token compression. Such discrete vocabularies provide a shared representation space by assigning visual features to a finite set of prototypes. Rather than using the codebook as a discrete tokenizer for replacing continuous visual features, our method treats the learned prototypes as a reusable assignment prior for grouping visual tokens across videos. The compressed tokens remain continuous aggregations of the original visual features, enabling offline-learned grouping structures to be efficiently reused through lightweight online assignment.

\section{Method}

\begin{figure*}[t]
    \centering
    \includegraphics[width=\textwidth]{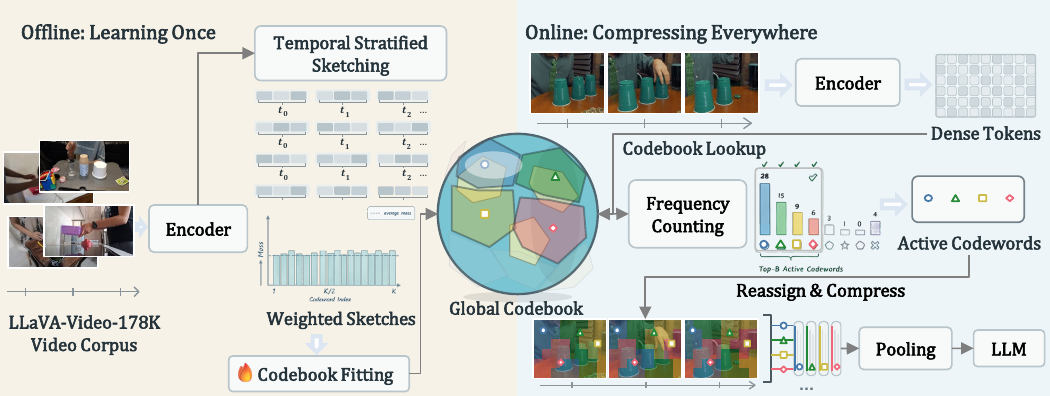}
    \caption{Overview of ONCE. (a) Offline, ONCE fits a global codebook from
    temporally stratified, source-count-weighted corpus sketches. (b) At
    inference, the fixed codebook selects groups supported by the current
    video, all projected visual tokens are reassigned, and each group is
    represented by the mean of its source embeddings.}
    \label{fig:pipeline}
\end{figure*}

\subsection{ONCE Framework Overview}
ONCE is an offline-to-online visual token compression framework that learns a
reusable global grouping prior once and applies it to arbitrary videos through
efficient token assignment and aggregation.

% Given a video $V$, a frozen visual encoder $E_{\mathrm{vis}}$ and the target
% model's visual projection or merger $P_{\mathrm{vis}}$ produce $N$ visual
% tokens of dimension $D$. ONCE replaces these tokens with a shorter sequence:
% \begin{equation}
% \begin{aligned}
% X&=P_{\mathrm{vis}}(E_{\mathrm{vis}}(V))
% \in\mathbb{R}^{N\times D},\\
% Z&=\operatorname{ONCE}_{C}(X;B)
% \in\mathbb{R}^{M\times D},
% \qquad M\leq\min(N,B).
% \end{aligned}
% \label{eq:visual_interface}
% \end{equation}

Given a video $V$, the frozen visual encoder $E_{\mathrm{vis}}$ and visual
projection/merger $P_{\mathrm{vis}}$ produce dense visual tokens:
\begin{equation}
X=P_{\mathrm{vis}}(E_{\mathrm{vis}}(V))
\in\mathbb{R}^{N\times D}.
\end{equation}

Instead of performing video-specific compression at inference time, ONCE
first learns a global visual codebook
$C=[c_1,\ldots,c_K]^{\mathsf T}$
offline from large-scale video data. During inference, the fixed codebook is
used as a grouping prior to assign dense tokens and aggregate them into a
compact sequence:
\begin{equation}
Z=\operatorname{ONCE}(X,C,B)
\in\mathbb{R}^{M\times D},
\quad M\leq\min(N,B).
\end{equation}
Here, $B$ bounds the number of pooled visual-content tokens, excluding
model-specific special and layout tokens, and $M$ is the input-dependent pooled
output length. We write
$\nu(v)=v/\max(\lVert v\rVert_2,\delta)$, with $\delta>0$, for
$\ell_2$ normalization in the cosine assignments below.

Importantly, the compressed tokens remain continuous aggregations of the
original visual embeddings rather than discrete code indices. Therefore,
ONCE requires no optimization for each input video and can efficiently
compress videos within the same frozen visual feature space.

% Here $C=[c_1,\ldots,c_K]^{\mathsf T}$ is a $K$-word codebook and $B$ bounds the
% pooled visual-content tokens, excluding boundary and layout tokens. We rebuild
% the model-specific attention mask and positions for $Z$. ONCE therefore shortens
% the language-model input but does not reduce visual encoding.

% Let $\nu(v)=v/\max(\lVert v\rVert_2,\delta)$, with $\delta>0$. Offline, ONCE
% summarizes training videos and fits $C$ in their frozen feature space. Online, the fixed codebook groups a new video's tokens, but each output averages that
% video's continuous embeddings. The codebook is reusable only within the same
% frozen encoder and projection or merger.

\subsection{Global Codebook Learning}
ONCE learns a global codebook from large-scale video data through two stages:
temporally aware corpus sketching and balanced prototype fitting.

\subsubsection{Temporally Stratified Corpus Sketch}
To learn a corpus-level codebook efficiently, we first construct a compact yet representative sketch from the training corpus. Instead of uniform sampling, we preserve temporal diversity by stratifying each video along its timeline.

We divide each video's normalized timeline into $L$ equal bins, assigning
$\tau=1$ to the final bin. For $N$ tokens, the sketch size is
$m=\min\{N,\operatorname{clip}(\lceil N/R\rceil,m_{\min},m_{\max})\}$, where
$R$ is the target number of source tokens per sketch token. We distribute the
$m$ representatives across nonempty bins as evenly as capacity permits, with
at least one per bin whenever $m$ is large enough.

Within each bin, spherical Lloyd ~\cite{spherical-lloyd} updates partition the normalized tokens into
local groups $A_j$. Each group stores a representative and the number of source
tokens it summarizes:
\begin{equation}
r_j=\nu\!\left(\sum_{i\in A_j}\nu(x_i)\right),
\qquad
w_j=|A_j|,
\qquad
\sum_j w_j=N.
\label{eq:weighted_representative}
\end{equation}
Distinct seeds avoid empty local groups. Aggregating all videos produces a corpus-level set of weighted sketches:
\[
\mathcal{D}=\{(r_u,w_u,g_u)\}_{u=1}^{U},
\]
where $g_u$ denotes the dataset category of the source video. Each sketch token
$r_u$ preserves its summarized visual content, while $w_u$ records the amount
of source-token mass it represents.

\subsubsection{Balanced Global Codebook Fitting}
Since naive clustering tends to allocate multiple prototypes to dominant visual patterns, we introduce balanced assignment to promote more even codeword utilization across the corpus.

For each representative $r_u$, we first define an effective weight
\begin{equation}
\omega_u=w_u p_{g_u}^{-\alpha},
\label{eq:representative_weight}
\end{equation}
where $w_u$ preserves the number of source tokens summarized by the
representative, $p_{g_u}$ denotes the frequency of its source category, and
$\alpha$ controls the strength of long-tail correction. When category labels
are unavailable, we set $\alpha=0$. This weighting scheme preserves the
contribution of high-mass representatives while preventing dominant categories
from overwhelming the learned codebook.
% For each valid category $g$, its representative frequency is
% $p_g=U^{-1}\sum_{u=1}^{U}\mathbf{1}[g_u=g]$. We assign representative $u$ the
% effective weight
% $\omega_u=w_u p_{g_u}^{-\alpha}$, using a neutral correction when the label is
% missing. Thus, $w_u$ preserves source-token mass and $\alpha$ controls the
% long-tail correction. Weighted spherical Lloyd updates initialize $K$
% codewords before balanced fitting.

% The matched controls in Table~\ref{tab:ablation_accuracy} separately compare
% the Full weighting scheme with $\alpha=0$, uniform representative weights
% ($\omega_u=1$), and hard spherical Lloyd in place of balanced Sinkhorn
% fitting.

Given a mini-batch of representatives $\mathcal{B}$, we normalize the
representative weights as row marginals and assign uniform mass to all
codewords:
\begin{equation}
a_u=\frac{\omega_u}{\sum_{v\in\mathcal{B}}\omega_v},
\qquad
b_k=\frac{1}{K}.
\label{eq:ot_marginal}
\end{equation}
The similarity between representative $r_u$ and codeword $c_k$ is measured by
cosine similarity:
\begin{equation}
S_{uk}=\nu(r_u)^{\mathsf T}\nu(c_k).
\label{eq:cosine_similarity}
\end{equation}

We formulate the codebook assignment as an entropy-regularized optimal
transport problem:
\begin{equation}
\begin{aligned}
\Gamma^\star
=&\arg\max_{\Gamma\geq0}
\langle\Gamma,S\rangle+\varepsilon H(\Gamma),\\
&\text{s.t.}\quad
\Gamma\mathbf{1}=a,\qquad
\Gamma^{\mathsf T}\mathbf{1}=b ,
\end{aligned}
\label{eq:balanced_ot}
\end{equation}
where $H(\Gamma)=-\sum_{u,k}\Gamma_{uk}\log\Gamma_{uk}$.
The row marginal preserves representative importance, while the uniform
column marginal encourages balanced utilization of codewords.

We solve Eq.~\ref{eq:balanced_ot} using log-domain Sinkhorn iterations
\cite{sinkhorn}. During optimization, straight-through
Gumbel--Softmax assignments
\cite{gumbel} enable differentiable codebook updates with
weighted reconstruction and code-usage objectives. After offline training,
the learned codebook $C$ is fixed and reused during inference.

% For a mini-batch $\mathcal{B}$, define normalized row weights $a_u$, uniform
% codeword weights $b_k$, and cosine similarities $S_{uk}$ as follows:
% \begin{equation}
% \begin{gathered}
% a_u=\frac{\omega_u}{\sum_{v\in\mathcal{B}}\omega_v},
% \qquad b_k=\frac{1}{K},
% \qquad S_{uk}=\nu(r_u)^{\mathsf T}\nu(c_k),\\
% \Gamma^\star=\argmax_{\Gamma\geq0}
% \left[\langle\Gamma,S\rangle+\varepsilon H(\Gamma)\right]
% \quad\text{s.t.}\quad
% \Gamma\mathbf{1}=a,\ \Gamma^{\mathsf T}\mathbf{1}=b .
% \end{gathered}
% \label{eq:balanced_ot}
% \end{equation}
% Here $H(\Gamma)=-\sum_{u,k}\Gamma_{uk}\log\Gamma_{uk}$ and $\varepsilon$ is the
% entropy coefficient. The exact row marginal preserves effective training mass,
% whereas the uniform column marginal balances mass across codewords. Individual
% assignments remain fractional.

% Log-domain Sinkhorn~\cite{sinkhorn} iterations approximate $\Gamma^\star$. Straight-through Gumbel--Softmax ~\cite{gumbel} assignments update $C$ with weighted reconstruction and
% code-usage losses, using $\omega_u$ in both transport and training. All model
% components remain frozen, and the fitted codebook is fixed at evaluation.

\subsection{Inference-Time Token Compression}
Given the offline-learned global codebook, ONCE performs inference-time
compression by using it as a reusable grouping prior. Dense visual tokens are
matched to codewords, while the current video determines which part of that
global geometry is active.

\subsubsection{Active Codeword Selection}

For a new video, ONCE first performs codebook lookup by assigning each visual
token to its nearest codeword in the frozen global codebook:
\begin{equation}
\widehat{k}_i=
\argmax_{k\in[K]}
\nu(x_i)^{\mathsf T}\nu(c_k),
\qquad
n_k=\sum_i\mathbf{1}[\widehat{k}_i=k].
\label{eq:initial_assignment}
\end{equation}

The initial assignment compares each input token against the reusable global
codebook and identifies the codewords activated by the current video. The
frequency $n_k$ records how many visual tokens are associated with codeword
$k$, reflecting how frequently the codeword represents visual tokens in the
input sequence. Based on these frequencies, ONCE selects the active codeword
set $\mathcal{K}_B$ by retaining the codewords with the largest nonzero
frequencies, where $|\mathcal{K}_B|\leq\min(B,K)$.

This frequency-based selection adapts the corpus-level codebook to each
individual video by removing inactive codewords before compression. The
selected active codewords form a compact grouping set for the subsequent
reassignment and token pooling stage.

\subsubsection{Reassignment and Token Pooling}

After selecting the active codewords, ONCE performs a second assignment over
the retained set $\mathcal{K}_B$ to construct the final grouping structure.
Specifically, each source token is reassigned to its nearest active codeword:
\begin{equation}
\begin{aligned}
k_i&=\argmax_{k\in\mathcal{K}_B}
\nu(x_i)^{\mathsf T}\nu(c_k),\\
z_k&=\frac{1}{|\mathcal{G}_k|}
\sum_{i\in\mathcal{G}_k}x_i .
\end{aligned}
\label{eq:complete_reassignment}
\end{equation}
where $\mathcal{G}_k$ denotes the set of source tokens assigned to active
codeword $k$.

The reassignment step updates token-to-codeword associations after removing
inactive codewords, ensuring that every source token is assigned to one of the
retained groups. Each group $\mathcal{G}_k$ is then compressed into a single
visual token $z_k$ through mean pooling over its original continuous
embeddings. Although the codebook determines the grouping structure, the
pooled representation is computed from the input video features rather than
the codeword itself.

The output tokens are ordered according to the frequency ranking of their
corresponding active codewords. Since the groups form a partition of all
source tokens, no visual token is discarded during pooling, and the output
sequence length satisfies $M\leq\min(N,B)$.

\paragraph{Inference complexity.}
The initial codebook lookup costs $O(NKD)$, while reassignment over the
selected codewords costs $O(N|\mathcal{K}_B|D)$. Frequency counting, partial
top-$B$ selection, and pooling cost $O(N+K\log B+ND)$. Since
$|\mathcal{K}_B|\leq B$, these additional terms are lower order when
$K\gg B$ and $ND\gg\log B$. In particular, reassignment uses at most a
fraction $B/K$ of the similarity computations required by the initial lookup.
Under these conditions, the total online compression overhead simplifies to
$O(NKD)$.

\section{Experiments}

% Generated by scripts/figures/plot_ergc_codebook_stats.py
\newcommand{\CbsOodN}{1000}
\newcommand{\CbsCrossPairN}{500}
\newcommand{\CbsOodFullR}{48.78}
\newcommand{\CbsOodRandomDelta}{1.70}
\newcommand{\CbsOodRandomCILow}{1.58}
\newcommand{\CbsOodRandomCIHigh}{1.85}
\newcommand{\CbsOodRandomReductionPct}{3.4}
\newcommand{\CbsOodHardDelta}{-0.61}
\newcommand{\CbsOodHardCILow}{-0.66}
\newcommand{\CbsOodHardCIHigh}{-0.57}
\newcommand{\CbsOodHardAdvantage}{0.61}
\newcommand{\CbsOodFullGain}{1.70}
\newcommand{\CbsOodFullGainCILow}{1.58}
\newcommand{\CbsOodFullGainCIHigh}{1.85}
\newcommand{\CbsOodFullPositive}{1000}
\newcommand{\CbsOodHardGain}{2.31}
\newcommand{\CbsOodHardGainCILow}{2.17}
\newcommand{\CbsOodHardGainCIHigh}{2.48}
\newcommand{\CbsOodHardPositive}{1000}
\expandafter\newcommand\csname CbsCrossDelta64\endcsname{3.32}
\expandafter\newcommand\csname CbsCrossCILow64\endcsname{3.02}
\expandafter\newcommand\csname CbsCrossCIHigh64\endcsname{3.62}
\expandafter\newcommand\csname CbsCrossReductionPct64\endcsname{4.8}
\expandafter\newcommand\csname CbsCrossPositivePairs64\endcsname{452}
\expandafter\newcommand\csname CbsCrossDelta128\endcsname{3.67}
\expandafter\newcommand\csname CbsCrossCILow128\endcsname{3.37}
\expandafter\newcommand\csname CbsCrossCIHigh128\endcsname{3.97}
\expandafter\newcommand\csname CbsCrossReductionPct128\endcsname{6.4}
\expandafter\newcommand\csname CbsCrossPositivePairs128\endcsname{480}
\expandafter\newcommand\csname CbsCrossDelta256\endcsname{3.52}
\expandafter\newcommand\csname CbsCrossCILow256\endcsname{3.22}
\expandafter\newcommand\csname CbsCrossCIHigh256\endcsname{3.82}
\expandafter\newcommand\csname CbsCrossReductionPct256\endcsname{7.1}
\expandafter\newcommand\csname CbsCrossPositivePairs256\endcsname{488}
\expandafter\newcommand\csname CbsCrossDelta512\endcsname{2.91}
\expandafter\newcommand\csname CbsCrossCILow512\endcsname{2.61}
\expandafter\newcommand\csname CbsCrossCIHigh512\endcsname{3.21}
\expandafter\newcommand\csname CbsCrossReductionPct512\endcsname{6.7}
\expandafter\newcommand\csname CbsCrossPositivePairs512\endcsname{492}
\newcommand{\CbsCrossDeltaMain}{3.67}
\newcommand{\CbsCrossCILowMain}{3.37}
\newcommand{\CbsCrossCIHighMain}{3.97}
\newcommand{\CbsCrossReductionPctMain}{6.4}
\newcommand{\CbsCrossPositivePairsMain}{480}
\newcommand{\CbsCrossMeanDeltaMain}{1.44}
\newcommand{\CbsCrossMeanCILowMain}{1.34}
\newcommand{\CbsCrossMeanCIHighMain}{1.55}
\newcommand{\CbsCrossRetainedGainMain}{14.3}
\newcommand{\CbsCrossRetainedCILowMain}{13.4}
\newcommand{\CbsCrossRetainedCIHighMain}{15.2}

% Auto-generated from validated training and inference ablations.
\newcommand{\AblUniformMV}{52.35}
\newcommand{\AblUniformLVB}{49.07}
\newcommand{\AblStaticRandomMV}{53.85}
\newcommand{\AblStaticRandomLVB}{51.68}
\newcommand{\AblStaticRandomBTwoFiveSixMV}{52.33}
\newcommand{\AblStaticRandomBTwoFiveSixLVB}{50.79}
\newcommand{\AblCodewordMV}{38.38}
\newcommand{\AblCodewordLVB}{41.74}
\newcommand{\AblMedoidMV}{54.03}
\newcommand{\AblMedoidLVB}{53.78}
\newcommand{\AblRandomGroupingMV}{45.33}
\newcommand{\AblRandomGroupingLVB}{46.00}
\newcommand{\AblStaticRandomUtilMV}{60.6}
\newcommand{\AblStaticRandomUtilLVB}{83.0}
\newcommand{\AblFullGroupUtilMV}{100.0}
\newcommand{\AblFullGroupUtilLVB}{100.0}
\newcommand{\AblMedoidFidelityMV}{0.944}
\newcommand{\AblMedoidFidelityLVB}{0.924}
\newcommand{\AblCodewordFidelityMV}{0.864}
\newcommand{\AblCodewordFidelityLVB}{0.887}
\newcommand{\AblMatchedFullMV}{51.45}
\newcommand{\AblMatchedFullLVB}{47.87}
\newcommand{\AblNoLongtailMV}{51.78}
\newcommand{\AblNoLongtailLVB}{48.99}
\newcommand{\AblHardLloydMV}{51.90}
\newcommand{\AblHardLloydLVB}{48.17}
\newcommand{\AblNoReassignMV}{53.83}
\newcommand{\AblNoReassignLVB}{53.48}
\newcommand{\AblNoReassignCoverageMV}{85.8}
\newcommand{\AblNoReassignCoverageLVB}{69.9}
\newcommand{\AblFramewiseMV}{54.08}
\newcommand{\AblFramewiseLVB}{52.80}
\newcommand{\AblationResultSummary}{Replacing pooled source embeddings with codeword prototypes reduces accuracy to 38.38\% on MVBench and 41.74\% on LongVideoBench. Skipping complete reassignment gives 53.83\% and 53.48\%, while retaining 85.8\% and 69.9\% of source tokens.}

\begin{table*}[!t]
\centering
\small
\setlength{\tabcolsep}{3.8pt}
\begin{tabular}{@{}clrrrr@{}}
\toprule
Token budget & Method & MVBench & EgoSchema & Video-MME & LongVideoBench \\
\midrule
23329 & LLaVA-OV-7B & 55.78 & 67.00 & 57.37 & 55.72 \\
\midrule
\multirow{3}{*}{32}
 & ForestPrune & 44.85 & 47.80 & 45.93 & 46.15 \\
 & FastVID & 40.67 & 39.60 & 43.85 & 43.08 \\
 & \textbf{ONCE (Ours)} & \textbf{47.93} & \textbf{55.20} & \textbf{48.44} & \textbf{47.49} \\
\midrule
\multirow{3}{*}{64}
 & ForestPrune & 47.67 & 52.80 & 47.96 & 46.45 \\
 & FastVID & 45.06 & 49.20 & 46.41 & 44.95 \\
 & \textbf{ONCE (Ours)} & \textbf{50.58} & \textbf{57.00} & \textbf{50.30} & \textbf{47.64} \\
\midrule
\multirow{3}{*}{128}
 & ForestPrune & 49.33 & 54.40 & 50.33 & 47.94 \\
 & FastVID & 49.32 & 54.80 & 49.44 & 47.42 \\
 & \textbf{ONCE (Ours)} & \textbf{52.20} & \textbf{60.40} & \textbf{51.07} & \textbf{49.89} \\
\midrule
\multirow{3}{*}{256}
 & ForestPrune & 52.80 & 57.20 & 52.22 & 50.26 \\
 & FastVID & 53.78 & 60.40 & \textbf{53.67} & \textbf{52.36} \\
 & \textbf{ONCE (Ours)} & \textbf{54.13} & \textbf{62.80} & 52.30 & 51.61 \\
\midrule
\multirow{3}{*}{512}
 & ForestPrune & 55.75 & 61.20 & 53.56 & 53.55 \\
 & FastVID & \textbf{56.45} & 59.80 & \textbf{56.56} & \textbf{54.75} \\
 & \textbf{ONCE (Ours)} & 54.50 & \textbf{62.40} & 53.26 & 52.95 \\
\midrule
\multirow{3}{*}{1024}
 & ForestPrune & \textbf{57.83} & 62.40 & 56.15 & \textbf{56.62} \\
 & FastVID & 57.65 & 59.40 & \textbf{57.26} & 56.17 \\
 & \textbf{ONCE (Ours)} & 54.28 & \textbf{62.60} & 53.85 & 54.00 \\
\bottomrule
\end{tabular}
\caption{LLaVA-OneVision-7B accuracy (\%) across visual-token budgets. ONCE
selects codewords jointly across all sampled frames of each video. Bold marks
the best compressed result within each budget.}
\label{tab:main_7b}
\end{table*}

\subsection{Experimental Setup}

\paragraph{Models, data, and baselines.}
We evaluate ONCE with the frozen LLaVA-OneVision-7B backbone~\cite{llava-onevision} on
MVBench~\cite{mvbench}, EgoSchema~\cite{egoschema},
Video-MME~\cite{video-mme}, and LongVideoBench~\cite{longvideobench},
which collectively span short-form temporal reasoning, long-video
comprehension, and broad multimodal video evaluation. We also instantiate
ONCE with Qwen3.5-9B~\cite{qwen35} on MVBench and LongVideoBench to evaluate its
generalization across video-language model architectures. For each backbone,
we learn a global
codebook once in its frozen visual feature space using the 0--30\,s subset
of LLaVA-Video-178K~\cite{llava-video}. The learned codebook is fixed and
reused for all downstream videos processed by the same backbone. Since
different backbones produce visual features in different representation
spaces, each codebook is dimension-matched to its corresponding backbone
while retaining the same ONCE design.

For LLaVA-OneVision-7B, we compare ONCE with two representative recent
training-free video token compression methods, ForestPrune ~\cite{forestprune} and FastVID ~\cite{fastvid}, under the same
backbone across different token budgets. ForestPrune performs high-ratio visual token pruning via
spatio-temporal token modeling, while FastVID adopts dynamic density-based pruning to reduce redundant visual tokens. These methods provide strong baselines for efficient video understanding by reducing redundant visual tokens. Supplementary experiments with LLaVA-OneVision-0.5B further evaluate
the scalability of ONCE across different model sizes.

\paragraph{Protocols and metrics.}
We evaluate video understanding performance under different token budgets
$B$ by varying the number of pooled visual tokens retained by ONCE.
Accuracy comparisons are conducted on MVBench~\cite{mvbench},
EgoSchema~\cite{egoschema}, Video-MME~\cite{video-mme}, and
LongVideoBench~\cite{longvideobench}, following the corresponding
evaluation protocols.

For efficiency evaluation, we profile inference cost under a fixed operating
point ($B=512$). All profiling experiments are conducted on a single NVIDIA
GeForce RTX~4090. We report four deployment-oriented metrics, including
visual token count, FLOPs, end-to-end latency, and peak GPU memory.

\paragraph{Implementation details.}
For each backbone, we learn a global codebook once in its frozen visual
feature space using the 0--30\,s subset of LLaVA-Video-178K~\cite{llava-video}.
The learned codebook is fixed and reused for all downstream videos
processed by the same backbone. Since different backbones produce visual
features in different representation spaces, we construct separate
codebooks for different backbones while preserving the same ONCE framework.

Unless otherwise specified, we use a codebook size of $K=8192$ and perform
online assignment with cosine similarity between visual tokens and
codewords. Additional details on dataset splits, video sampling, decoding,
and profiling settings are provided in the supplementary material.

\subsection{Main Results}

\paragraph{LLaVA-OneVision-7B.}
At $B\in\{32,64,128\}$, ONCE attains the highest accuracy among compressed
methods on all four benchmarks (Table~\ref{tab:main_7b}), indicating that the
reusable grouping prior is most valuable when token capacity is scarce. As
$B$ grows, pruning baselines close or reverse the gap; the result is therefore
low-budget robustness rather than uniform dominance.

\begin{table}[!htbp]
\centering
\small
\setlength{\tabcolsep}{4.0pt}
\begin{tabular}{@{}clrr@{}}
\toprule
Token budget & Method & MV & LVB \\
\midrule
Standard & Qwen3.5-9B & 70.18 & 65.37 \\
\midrule
128  & ONCE & 56.23 & 51.46 \\
256  & ONCE & 58.90 & 53.78 \\
512  & ONCE & \textbf{63.25} & \textbf{56.40} \\
1024 & ONCE & 61.95 & 54.43 \\
\bottomrule
\end{tabular}
\caption{Qwen3.5-9B accuracy (\%) across visual-token budgets. Videos are
sampled at 2 frames per second (FPS), with at most 256 frames.}
\label{tab:main_operating_point}
\end{table}

\paragraph{Qwen3.5-9B.}
ONCE transfers to Qwen3.5-9B, with $B=512$ best on both benchmarks
(Table~\ref{tab:main_operating_point}). The trend suggests that
downstream performance under codebook compression need not improve linearly
with the retained-token budget. Thus, $B$ should be selected empirically rather
than treated as a direct proxy for representation quality. Unlike LLaVA's
tight-budget advantage and mixed higher-budget rankings, Qwen provides a
complementary budget trend. All Qwen rows reuse one codebook without
retraining.

\paragraph{Efficiency.}
ONCE removes most visual-token prefill work on both backbones, yet end-to-end
speedup is smaller and varies by workload (Table~\ref{tab:main_efficiency}).
Compression occurs after visual encoding, leaving encoder and decoding costs
unchanged; the remaining runtime therefore limits realized acceleration. The
larger LLaVA speedups indicate that deployment gains depend on the backbone's
prefill share rather than token reduction alone. Memory falls more modestly
because model weights and non-visual state remain resident.

\begin{table}[!htbp]
\centering
\small
\setlength{\tabcolsep}{2.0pt}
\begin{tabular}{@{}llrrrrr@{}}
\toprule
Model & Data & Vis. & Tok.$\downarrow$ & FLOPs$\downarrow$ & E2E$\uparrow$ & Mem.$\downarrow$ \\
 & & (avg.) & (\%) & (\%) & ($\times$) & (\%) \\
\midrule
LLaVA & MV  & 467.70 & 98.00 & 98.29 & 4.49 & 16.36 \\
LLaVA & LVB & 503.64 & 97.84 & 97.95 & 3.35 & 16.69 \\
Qwen  & MV  & 340.77 & 93.48 & 91.22 & 1.79 & 15.58 \\
Qwen  & LVB & 701.31 & 98.82 & 98.04 & 2.22 & 14.39 \\
\bottomrule
\end{tabular}
\caption{Efficiency at $B=512$; arrowed columns are relative to matched
uncompressed inference.}
\label{tab:main_efficiency}
\end{table}
\FloatBarrier

\subsection{Ablation Study}
Table~\ref{tab:ablation_accuracy} separates the roles of temporal
partitioning, reassignment, grouping, and output representation. All
inference-side variants reuse Full's codebook.

\begin{table}[!htbp]
\centering
\small
\setlength{\tabcolsep}{3.0pt}
\begin{tabular}{@{}llrr@{}}
\toprule
Variant & Changed component & MV & LVB \\
\midrule
Full (Table~\ref{tab:main_7b}) & -- & 54.50 & 52.95 \\
Single-bin & Partition & 39.03 & 41.96 \\
w/o complete reassignment & Reassignment & \AblNoReassignMV & \AblNoReassignLVB \\
Random grouping & Grouping & \AblRandomGroupingMV & \AblRandomGroupingLVB \\
Codeword output & Representation & \AblCodewordMV & \AblCodewordLVB \\
\bottomrule
\end{tabular}
\caption{LLaVA-OneVision-7B ablations at $B=512$; Random grouping preserves
Full's group sizes and randomizes membership.}
\label{tab:ablation_accuracy}
\end{table}

Single-bin training and codeword output produce the largest losses even though
they alter different stages of ONCE. The two failures are complementary.
Single-bin training removes temporal stratification from the offline corpus
sketch, whereas codeword output substitutes fixed prototypes for current-video
source means. Their similar degradation supports a division of labor: offline
learning organizes the feature space, while online pooling retains
instance-specific evidence. The codebook should therefore guide group
membership rather than serve as the compressed representation.

Random grouping keeps Full's group-size profile and source-mean output rule
while breaking feature-based membership. Its lower accuracy on both benchmarks
indicates that group size alone does not account for Full's result; codebook
membership carries task-relevant structure. Removing complete reassignment
changes accuracy much less, although the supplementary coverage diagnostic
shows that some source tokens stop contributing. Reassignment therefore
guarantees complete representation, while grouping and output construction
determine what the compressed sequence retains.
% Having identified how the codebook should be used, we next examine whether the selected codebook is broadly utilized and whether its geometry remains useful beyond the training cache.

\subsection{Codebook Analysis}
Beyond downstream accuracy, we analyze whether the offline-learned
codebook captures reusable visual grouping structures in the feature space, rather than merely storing training examples. We analyze the selected Full codebook in two stages: utilization on the
training corpus and coverage of public-video features. The training
diagnostics in Table~\ref{tab:training_codebook_diagnostics} describe three
complementary properties. Active codes measure occupancy, entropy-effective
capacity discounts highly uneven assignment mass, and mean cosine error
measures how closely cached features match their nearest codewords. Together,
they distinguish a codebook that is merely nonempty from one whose capacity is
distributed across the corpus.

\begin{table}[!htbp]
\centering
\small
\setlength{\tabcolsep}{5.2pt}
\begin{tabular}{@{}lcc@{}}
\toprule
Training-corpus diagnostic & Estimate & 95\% CI \\
\midrule
Active codes (\%) & 99.93 & -- \\
Entropy-effective capacity (\%) & 53.39 & [53.25, 53.42] \\
Mean cosine error (\(\times10^{-3}\)) & 4.763 & [4.750, 4.775] \\
\bottomrule
\end{tabular}
\caption{Full-codebook utilization on the Stage-A cache ($K=8192$). The 95\%
confidence intervals (CIs) use video-level bootstrap resampling.}
\label{tab:training_codebook_diagnostics}
\end{table}

\begin{figure}[!ht]
\centering
\includegraphics[width=\columnwidth]{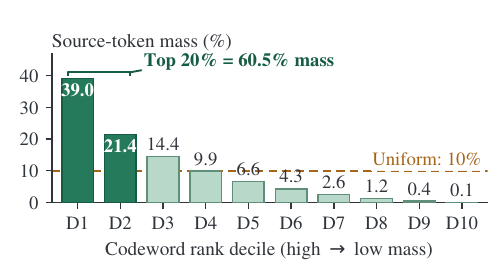}
\caption{Training-corpus codebook utilization. Nearly all codewords are
active, although assignment mass remains concentrated.}
\label{fig:training_codebook_usage}
\end{figure}

The table establishes broad occupancy but less uniform effective use. To
localize this imbalance, Figure~\ref{fig:training_codebook_usage} orders
codewords by assigned source-token mass and aggregates them into equal-count
deciles. Bar height represents each decile's share of the assignments; the
emphasized bars isolate the highest-ranked fifth, and the dashed line provides
the uniform reference.

Figure~\ref{fig:training_codebook_usage} shows that assignment mass is
concentrated in the leading deciles and then tapers across the remaining
codebook. Thus, near-complete activation does not imply balanced use: common
visual structures carry much of the workload, while lower-frequency regions
still retain dedicated codewords.

\begin{figure}[!ht]
\centering
\includegraphics[width=\columnwidth]{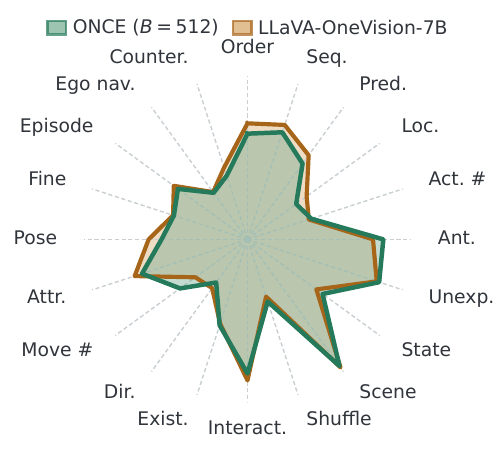}
\caption{MVBench subtask performance. ONCE remains competitive across diverse
video-reasoning tasks, with its clearest gains in event counting and change
recognition.}
\label{fig:mvbench_task_radar}
\end{figure}

Corpus utilization does not show how a fixed token budget redistributes
accuracy across video capabilities.
\noindent\begin{minipage}{\columnwidth}
\hspace*{1em}%
We therefore expand the $B=512$ MVBench comparison in
Figure~\ref{fig:mvbench_task_radar}. Every spoke uses the same raw scale, so an
outward separation directly indicates higher accuracy on that task.
\end{minipage}

\FloatBarrier

Figure~\ref{fig:mvbench_task_radar} shows a selective rather than uniform
retention pattern. ONCE is higher on seven tasks, with the clearest outward
shifts in moving count, action antonym, state change, and object shuffle. The
larger inward shifts occur in action localization, fine-grained pose, action
prediction, character order, and counterfactual inference. Grouped by
capability, the pattern suggests that global grouping retains event-level
counts and state transitions more reliably than precise temporal alignment or
subtle pose cues.

A supplementary public-video stress test finds lower mean and tail cosine
residuals for Full than for equally sized random-exemplar codebooks. This result
suggests that the fitted geometry remains useful beyond the training cache.

Together with the ablation, these diagnostics suggest a two-part mechanism.
The global codebook supplies reusable partition geometry, while the current
video determines which regions are active and provides the content of the
compressed tokens. This division explains how ONCE can reuse an offline
structure without making every video share the same compressed
representation, and it connects the codebook analysis to the accuracy and
efficiency gains reported above.
\FloatBarrier

\section{Conclusion}

Efficient long-video inference requires compressing dense visual-token
sequences without optimizing a new compression strategy for every input. ONCE
learns offline a global visual codebook that captures corpus-level statistics
in the frozen feature space of the target Video-LLM. At inference, the fixed
codebook is reused for codebook lookup, complete reassignment, and token
aggregation. The codebook determines the grouping structure, while online mean
pooling preserves task-relevant continuous representations from the current
video.

Across the tested token budgets and Video-LLM backbones, ONCE maintains
competitive task performance. Under extreme token budgets
($B\in\{32,64,128\}$), ONCE achieves the highest accuracy among the compared
compression baselines on all four benchmarks. Across the four
backbone--dataset profiling configurations at $B=512$, visual-token prefill
FLOPs fall by $96.38\pm3.44$ percentage points, and end-to-end inference is
$(2.96\pm1.21)\times$ faster (mean $\pm$ standard deviation). Under our design
and configuration, $99.93\%$ of the codewords are active, although
source-token assignment mass remains concentrated among the leading
codewords. This pattern indicates a feature-space partition aligned with the
corpus statistics: frequent visual patterns receive more assignments, while
dedicated codewords still cover lower-frequency regions. Lower mean and tail
cosine residuals than random-exemplar codebooks on the public-video stress set
further support the use of this statistical structure beyond the training
cache. The ablation indicates that combining reusable global structure with
video-specific representation aggregation is an effective design for ONCE.

\paragraph{Limitations and future work.}
ONCE still requires a separate codebook for each target feature space.
The fixed codebook is learned from a finite offline sample of the frozen visual
feature space. Broader coverage of that space could improve the fitted grouping
prior, but constructing a sufficiently diverse sample is currently limited by
data and computational resources.
Compression occurs after visual encoding and therefore does not reduce
vision-encoder computation. The task profile also suggests weaker retention of
precise temporal localization, event ordering, and fine-grained pose cues than
of event counts and state changes. Future work could share or transfer
codebooks across related backbones and move compression earlier into visual
encoding. Although the current evaluations focus on video question answering,
adaptations to the target data, representation space, and downstream objective
may extend this compression approach to other tasks and application domains.

% Additional experiments are submitted separately in SupplementaryMaterial2027.tex.

% \section*{Acknowledgments}

\bibliography{aaai2027}
\newpage
\section{Additional Experiments}

\subsection{Evaluation Details}

\paragraph{Models and accuracy evaluation.}
We learn a separate ONCE codebook from LLaVA-Video-178K for each frozen visual
feature space. The main LLaVA-OneVision-7B evaluation uses MVBench, EgoSchema,
Video-MME, and LongVideoBench; the Qwen3.5-9B evaluation uses MVBench and the
visual-only validation split of LongVideoBench. All ONCE accuracy results use
whole-video codeword selection. The Qwen3.5-9B budget sweep uses BF16,
non-thinking greedy decoding, 2-FPS sampling, and at most 256 frames, with 4,000
MVBench and 1,337 LongVideoBench samples. The LLaVA-OneVision-0.5B results
below provide an additional model-scale comparison.

\paragraph{Codebook reuse across token budgets.}
For a fixed backbone and frozen visual feature space, every sweep over the
inference budget \(B\) reuses the same learned \(K=8192\) codebook; changing
\(B\) only changes how many active codewords are retained online and does not
trigger codebook retraining. Only an experiment that changes the offline
codebook-learning procedure or the backbone's visual feature space requires a
newly trained codebook.

\paragraph{Downstream reuse and corpus audit.}
ONCE learns a grouping prior from frozen visual features without using
downstream questions, answers, task labels, or evaluation predictions. We also
compared Stage-A source identities with the downstream records available to the
evaluation pipeline. The source annotations contain 94,583 unique media paths,
of which 94,581 were successfully encoded. MVBench has no source-qualified
logical-path match with these records. Its 41 shared generic basenames arise
from the naming conventions of PerceptionTest and CLEVRER rather than a shared
source. None of the 753 unique LongVideoBench validation identifiers matches a
normalized training identifier. These results find no direct identity reuse in
the audited records. More generally, cross-video reuse in ONCE means that one
corpus-level codebook is applied without task-specific or per-video fitting;
the mechanism does not rely on downstream supervision.

\paragraph{Qwen evaluation alignment.}
The Standard and Full \(B=512\) Qwen3.5-9B results use the same benchmark
splits, sample sets, BF16 non-thinking greedy decoding, 2-FPS input, and
at-most-256-frame cap. These are the settings reported in main Table~2.

\paragraph{Checkpoint size and indirect verification.}
The full-precision LLaVA-OneVision-7B checkpoint is 117,442,176 bytes
(112.0 MiB), and the Qwen3.5-9B checkpoint is 134,272,832 bytes
(128.1 MiB). Lossless ZIP compression reduces them only to 108,806,213 bytes
(103.8 MiB) and 124,634,011 bytes (118.9 MiB), respectively. Thus, either
checkpoint alone exceeds the 50 MB code-and-data upload limit, and the binary
weights are omitted from the submitted archive. For indirect verification, the
archive records each artifact and codebook-tensor SHA-256, the expected
\(8192\times3584\) and \(8192\times4096\) shapes, and the pre-packaging results
of strict loading, finite-value checks, nearest-code assignment, and the
model-specific temporal-compression paths. It also contains the fixed
configurations and scripts needed to retrain and locally validate either
checkpoint. Together, these records define an indirect verification path for
artifact identity and the tested inference interfaces under the submission
size limit.

\paragraph{Random seeds and reproducibility.}
The reported codebook-training, accuracy, control, and profiling jobs use
the fixed seed 42. The training-cache video-level bootstrap uses seed 20260729,
and deterministic public-video stress-set construction uses seed 20260731. The
five public-video random-exemplar codebooks use seeds 20260801--20260805. The
public-video coverage and cross-view analyses each use 5,000 resamples with
bootstrap seed 20260811. All seeds are explicit configuration values reused
across matched comparisons.

\paragraph{Matched efficiency evaluation.}
All matched profiling runs were conducted locally on one NVIDIA GeForce
RTX~4090 in a 48-GiB memory configuration (49,140 MiB reported device memory),
with batch size 1. Each ONCE run is paired with standard inference without
visual-token compression using the same backbone, benchmark, sample set, and
input settings. LLaVA-OneVision videos are decoded at 1 FPS and uniformly
sampled to 32 frames. Qwen3.5 uses 2-FPS sampling with 4--256 frames. The main
efficiency comparison uses $B=512$. The reported visual-token count includes
pooled content tokens and model-specific special or layout tokens, while $B$
is an upper bound on pooled content tokens. We report prefill FLOPs because
ONCE directly changes the sequence entering the language model; the estimate
uses its actual input length and isolates the targeted computation.
End-to-end latency covers the complete path, including visual encoding, the
\(O(NKD)\) lookup, pooling, and autoregressive decoding. The two metrics
separate the direct prefill saving from the realized system effect.
Table~\ref{tab:absolute_peak_memory} reports the absolute PyTorch peak reserved
memory rather than peak allocated memory or \texttt{nvidia-smi} process usage.

\paragraph{One-time offline cost and storage.}
Table~\ref{tab:offline_cost} reports completed wall-clock measurements for one
recorded single-RTX~4090 run per backbone. Stage A extracts the frozen
representatives and writes the fitting cache; Stage B fits the \(K=8192\)
codebook from that cache.

\begin{center}
\centering
\footnotesize
\setlength{\tabcolsep}{2.0pt}
\begin{tabular}{@{}lrrrrr@{}}
\toprule
Backbone & Stage A & Stage B & Cache & Reps. & Codebook \\
\midrule
LLaVA-7B & 15.62 h & 4.67 min & 32.93 GiB & 4.918 M & 112 MiB \\
Qwen-9B  & 1.39 h  & 3.76 min & 15.15 GiB & 1.980 M & 128 MiB \\
\bottomrule
\end{tabular}
\captionof{table}{One-time codebook-construction cost and storage. Times are
completed single-run wall-clock measurements. Cache is fitting-time storage;
Codebook is the deployment checkpoint footprint.}
\label{tab:offline_cost}
\end{center}

Stage A dominates the measured one-time cost, while Stage B takes minutes;
neither is charged per evaluation video. The cache can be discarded after
validation and deployment retains only the codebook. For \(N\) downstream
inferences, the amortized one-time cost is
\((T_{\mathrm{A}}+T_{\mathrm{B}})/N\). Relative to Dense inference, a
configuration-specific break-even count is
\[
N^\star=\frac{T_{\mathrm{A}}+T_{\mathrm{B}}}
{\ell_{\mathrm{Dense}}-\ell_{\mathrm{ONCE}}}.
\]
For the LLaVA MVBench \(B=512\) measurements in
Table~\ref{tab:efficiency}, 56,508.54 seconds of offline work and a
2,634.43-ms per-video latency difference give \(N^\star\approx21{,}450\).
This calculation expresses the one-time construction cost at a concrete
deployment scale; the same equation gives the corresponding crossover for any
backbone, benchmark, and inference configuration.

\paragraph{Inference controls.}
The inference controls below use the frozen $B=512$ operating point from main
Table 1. Fixed-random selection and codeword output use the same codebook,
sample sets, and frozen inference runtime as the Full reference.
Fixed-random selection uses one shared subset of 512 codewords, followed by
complete reassignment. Codeword output retains the
Full assignments but emits the selected prototypes instead of source-embedding
means. Nearest-source output, referred to internally as medoid output, also
retains the Full assignments and emits the assigned source token with maximum
cosine similarity to each pooled mean.

\paragraph{Full codebook configuration.}
Table~\ref{tab:full_codebook_config} gives the single Full configuration used
for each backbone. These are the reference settings for all reported budget
sweeps. An inference-budget experiment changes only \(B\).

\begin{table*}[t]
\centering
\footnotesize
\setlength{\tabcolsep}{4pt}
\renewcommand{\arraystretch}{1.08}
\begin{tabular}{@{}p{0.18\textwidth}p{0.38\textwidth}p{0.38\textwidth}@{}}
\toprule
Setting & LLaVA-OneVision-7B Full & Qwen3.5-9B Full \\
\midrule
Codebook shape
  & \(8192\times3584\)
  & \(8192\times4096\) \\
Feature source
  & Projected patch-grid features
  & Visual-merger output \\
Training videos
  & 94,581 LLaVA-Video-178K videos
  & 94,581 LLaVA-Video-178K videos \\
Stage-A video sampling
  & 1 FPS; 32 uniformly sampled frames
  & 2 FPS; 4--2,048 frames \\
Accuracy evaluation
  & FP16; 1 FPS; 32 uniformly sampled frames
  & BF16; non-thinking greedy; 2 FPS; at most 256 frames \\
Adaptive sketch
  & \(R=256,\ m_{\min}=16,\ m_{\max}=52\) (128 requested)
  & \(R=256,\ m_{\min}=16,\ m_{\max}=128\) \\
Temporal sketch bins
  & 4
  & 4 \\
Cached representatives
  & 4,918,212 BF16 vectors
  & 1,980,488 BF16 vectors \\
Source tokens covered
  & 2,206,385,568
  & 436,884,976 \\
Representative weighting
  & Source count with long-tail exponent \(\alpha=0.5\)
  & Source count \\
Initialization
  & 65,536 candidates and 3 weighted spherical-Lloyd refinements
  & 8,192 sampled candidates without post-refinement \\
Sinkhorn
  & FP32; \(\varepsilon=0.05\); 5 normalization steps
  & FP32; \(\varepsilon=0.05\); 5 normalization steps \\
Optimization
  & AdamW; batch 65,536; LR \(10^{-3}\); WD \(10^{-2}\);
    cosine schedule; gradient clip 1
  & AdamW; batch 65,536; LR \(10^{-3}\); WD \(10^{-2}\);
    cosine schedule; gradient clip 1 \\
Regularization
  & Commitment 0.25; usage 0.01; temperature
    \(1.0\!\rightarrow\!0.01\), decay 0.999
  & Commitment 0.25; usage 0.01; temperature
    \(1.0\!\rightarrow\!0.01\), decay 0.999 \\
Checkpoint selection
  & At most 5 epochs; patience 2; refined initializer (epoch 0) selected
  & 5 epochs; patience 2; epoch 5 selected \\
Inference rule
  & Whole-video selection, complete reassignment, source-mean output
  & Whole-video selection, complete reassignment, source-mean output \\
Selection rounds \(Q\)
  & 4
  & 4 \\
Main operating point
  & \(B=512\)
  & \(B=512\) \\
Random seed
  & 42
  & 42 \\
\bottomrule
\end{tabular}
\caption{Full codebook-training and inference configuration. The 5,000,000
representative cap resolves the LLaVA per-video sketch maximum from the
requested 128 to 52. All token-budget experiments for a fixed backbone reuse
the corresponding Full codebook.}
\label{tab:full_codebook_config}
\end{table*}

Checkpoint selection treats the refined initializer as epoch 0 and subsequent
Sinkhorn epochs as additional candidates. It retained the refined initializer
for LLaVA and epoch 5 for Qwen. This backbone-specific selection is deliberate:
it retains the initialized geometry when adequate and later balanced updates
when they improve the selection criterion, while leaving the shared online
rule unchanged.

Across the two backbones, Full keeps \(K\), \(Q\), the main operating point,
and the online rule fixed, while the feature source and dimension follow the
visual encoder. This separates the shared ONCE design---whole-video selection,
complete reassignment, and source-mean output---from the codebook fitted in
each feature space. The cross-backbone results therefore test the same
compression mechanism with a backbone-matched grouping prior.

\paragraph{LLaVA ablation evaluation.}
The inference-policy, codebook-training, and output-representation controls
reported in the ablation tables use the frozen LLaVA-OneVision-7B model with
FP16 inference and 32 uniformly sampled frames. We evaluate all 4,000 MVBench
and 1,337 LongVideoBench samples. Unless the variant name changes one of these
steps, inference uses whole-video codeword selection, complete reassignment,
and source-embedding mean pooling at the displayed budget.

\paragraph{Public-video stress diagnostic.}
The coverage and support-origin diagnostics use the Full codebook; the coverage
control additionally uses matched Hard-Lloyd.
The diagnostic contains \CbsOodN{} unique Wikimedia Commons video file pages,
selected deterministically from the API video index. Eligible pages have a
free-license record, a 3--180\,s duration, and an official WebM derivative with
at least 360 pixels on the short side. We cap each normalized creator and
uploader at 30 retained pages; the realized set contains 367 normalized creator
clusters and 282 uploaders. File-page ids, downloaded hashes, and five-frame
perceptual signatures are all unique. The manifest records the file page,
source and derivative URLs, source-page SHA-1, downloaded-derivative SHA-256,
license, credit, upload timestamp, creator, uploader, and decoded metadata.
The downloaded media remain outside the manuscript repository.

Comparing file size and then SHA-256 against all 94,699 files in the audited
local Stage-A tree found no byte-identical derivative. We designed this set as
a public-video stress diagnostic rather than as a formal OOD claim. Its role is
to test feature coverage across independently indexed public file pages under
creator and uploader caps. Re-encodes, related events, semantic overlap, and
model-pretraining exposure are outside the definition of this diagnostic.

\begin{figure}[t]
\centering
\includegraphics[width=\columnwidth]{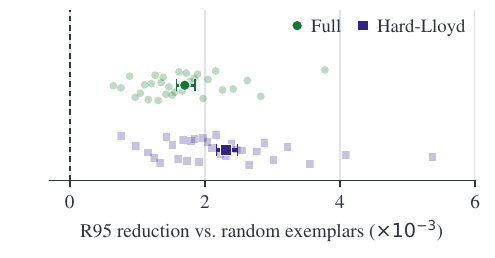}
\caption{Public-video tail-coverage gains over random exemplars. Small markers
are observed order statistics; enlarged markers and whiskers show full-sample
means and pointwise 95\% CIs. Positive favors learned codebooks.}
\label{fig:supp_codebook_coverage}
\end{figure}

Each video is uniformly sampled to 32 frames and encoded by the frozen
LLaVA-OneVision-7B vision tower and projector without spatial pooling, yielding
\(32\times729\) BF16 vectors of dimension 3,584. Feature identity is checked
against the Stage-A cache and every codebook checkpoint.
Figure~\ref{fig:supp_codebook_coverage} compares Full and matched Hard-Lloyd
with five \(K=8192\) random-exemplar codebooks.
Assignment follows the production BF16 order (cast, normalize, matrix
multiply). The primary metric is each video's 95th percentile FP32 cosine
residual, R95: it targets the least-covered 5\% of tokens while retaining about
1,166 tail tokens per video. Mean residual and R50/R90/R99 are sealed as
sensitivity measures; they do not replace R95 in this diagnostic.

For the supplementary support-origin diagnostic, even and odd frames form two
interleaved views. One view selects up to \(B\) supported Full-codebook ids;
the other is fully reassigned to that support and pooled using only its own
BF16 source means, after which the direction is reversed and averaged. Videos
are partitioned into
\CbsCrossPairN{} disjoint reciprocal pairs by projected-feature-centroid
similarity, requiring different creators and uploaders whenever an eligible
partner remains. This makes the other-video control visually conservative
rather than trivially cross-domain. The realized support is
\(\min(B,\text{number of supported ids})\), so \(B\) is an upper bound.
No realized pair shares a normalized creator or uploader, and the median
projected-centroid cosine is 0.998.

\begin{figure}[!ht]
\centering
\includegraphics[width=\columnwidth]{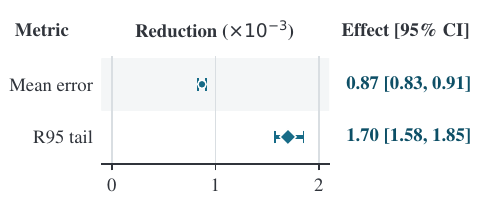}
\caption{Cross-video feature coverage. Full improves both typical- and
tail-token coverage over random exemplars.}
\label{fig:full_codebook_residual}
\end{figure}

\begin{figure}[t]
\centering
\includegraphics[width=\columnwidth]{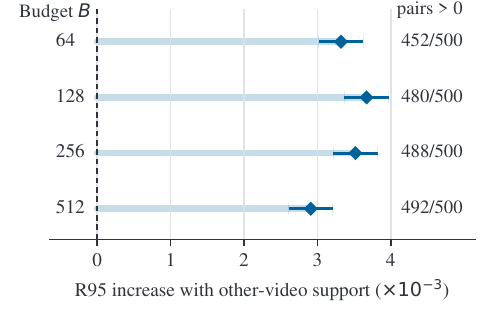}
\caption{Effect of support origin on pooled R95. Markers show
donor-minus-current means, whiskers show simultaneous 95\% CIs, and labels
count positive pairs. Positive favors current-video support.}
\label{fig:supp_cross_view_transfer}
\end{figure}

We use 5,000 nonparametric resamples. Videos and the five random-codebook seeds
are resampled for Figures~\ref{fig:full_codebook_residual}
and~\ref{fig:supp_codebook_coverage}; disjoint pairs are resampled for
Figure~\ref{fig:supp_cross_view_transfer}. For legibility, the small symbols
in Figure~\ref{fig:supp_codebook_coverage} are 30 observed order
statistics selected at evenly spaced empirical percentiles from 2.5 to 97.5;
they are not a 30-video subsample. All displayed means and intervals use the
full \CbsOodN{} videos. Figure~\ref{fig:supp_codebook_coverage} reports
pointwise percentile 95\% intervals. Because all four budgets reuse the same
pairs, the support-origin figure instead reports a non-studentized
max-deviation 95\% simultaneous band across the prespecified budgets; its
right labels count positive effects among all \CbsCrossPairN{} pairs. The
resampling treats videos or pairs as units; creator and uploader concentration
is handled by the construction caps rather than a second hierarchical model.

Figure~\ref{fig:full_codebook_residual} summarizes the paired
random-minus-Full effects for per-video mean residual and R95. Positive values
favor Full.

\FloatBarrier

Both estimates and their confidence intervals lie to the right of zero. The
larger tail reduction indicates that learned geometry is especially helpful
for poorly covered tokens, while the positive mean effect shows that the
improvement is not confined to the tail.

Full lowers nearest-codeword R95 by
\(\CbsOodFullGain\times10^{-3}\) (95\% CI
\([\CbsOodFullGainCILow,\CbsOodFullGainCIHigh]\times10^{-3}\)), or
\CbsOodRandomReductionPct\% of the random R95; both learned codebooks are
lower on all \CbsOodN{} videos. Hard-Lloyd-minus-Full is
\(\CbsOodHardDelta\times10^{-3}\) (95\% CI
\([\CbsOodHardCILow,\CbsOodHardCIHigh]\times10^{-3}\)). Full and Hard-Lloyd
therefore both improve coverage over random exemplars; optimizer-specific
ordering is separate from the learned-versus-random question measured here.

The learned-versus-random coverage gain remains positive for mean residual and
R50/R90/R99, so the result is not specific to R95.

The simultaneous-CI lower bounds in
Figure~\ref{fig:supp_cross_view_transfer} are positive at all four budgets.
At \(B=128\), donor support increases pooled R95 by
\(\CbsCrossDeltaMain\times10^{-3}\) (simultaneous 95\% CI
\([\CbsCrossCILowMain,\CbsCrossCIHighMain]\times10^{-3}\)); the effect is
positive for \CbsCrossPositivePairsMain{} of \CbsCrossPairN{} pairs. Donor
support also increases mean residual by
\(\CbsCrossMeanDeltaMain\times10^{-3}\) (pointwise 95\% CI
\([\CbsCrossMeanCILowMain,\CbsCrossMeanCIHighMain]\times10^{-3}\)), while
current-video support covers \CbsCrossRetainedGainMain{} percentage points
more of the target view's initial Full-code assignments (pointwise 95\% CI
[\CbsCrossRetainedCILowMain,\CbsCrossRetainedCIHighMain] points). These
sensitivity measures agree with the R95 diagnostic, but this secondary
comparison is not used as the main evidence for codebook quality.

Table~\ref{tab:appendix_inference_ablation} separates adaptive selection,
complete reassignment, and temporal selection scope. Full exceeds fixed-random
selection at both budgets on both benchmarks, which supports selecting a
video-conditioned subset rather than retaining one global subset for every
input. Complete reassignment restores full source-token coverage, while its
accuracy change is small and has opposite signs on MVBench and LongVideoBench.
Its primary role is therefore to ensure that every source token contributes;
selection determines which supported regions compete for the output budget.
Frame-wise selection remains close to Full, suggesting that whole-video
selection is a consistent default rather than the sole source of the
downstream result.

\begin{table}[t]
\centering
\small
\setlength{\tabcolsep}{3.8pt}
\begin{tabular}{@{}lrrrrr@{}}
\toprule
\multicolumn{2}{c}{} & \multicolumn{2}{c}{Accuracy (\%)} &
\multicolumn{2}{c}{Coverage (\%)} \\
\cmidrule(lr){3-4}\cmidrule(l){5-6}
Inference policy & $B$ & MV & LVB & MV & LVB \\
\midrule
Full                     & 256 & 54.13 & 51.61 & 100.0 & 100.0 \\
Fixed-random selection   & 256 & \AblStaticRandomBTwoFiveSixMV
                         & \AblStaticRandomBTwoFiveSixLVB & 100.0 & 100.0 \\
Full                     & 512 & 54.50 & 52.95 & 100.0 & 100.0 \\
Fixed-random selection   & 512 & \AblStaticRandomMV & \AblStaticRandomLVB
                         & 100.0 & 100.0 \\
\midrule
w/o complete reassignment & 512 & \AblNoReassignMV & \AblNoReassignLVB
                         & \AblNoReassignCoverageMV & \AblNoReassignCoverageLVB \\
Frame-wise selection     & 512 & \AblFramewiseMV & \AblFramewiseLVB
                         & 100.0 & 100.0 \\
\bottomrule
\end{tabular}
\caption{Selection and inference-policy ablations. Coverage is the fraction of
source tokens represented by the output groups. Fixed-random selection uses
the same codeword permutation at both budgets.}
\label{tab:appendix_inference_ablation}
\end{table}
\FloatBarrier

\begin{center}
\centering
\small
\setlength{\tabcolsep}{6.0pt}
\begin{tabular}{@{}lrr@{}}
\toprule
Stage-B variant & MV & LVB \\
\midrule
Weighted + long-tail + Sinkhorn & \AblMatchedFullMV & \AblMatchedFullLVB \\
Uniform weights          & \AblUniformMV & \AblUniformLVB \\
w/o long-tail            & \AblNoLongtailMV & \AblNoLongtailLVB \\
Hard-Lloyd               & \AblHardLloydMV & \AblHardLloydLVB \\
\bottomrule
\end{tabular}
\captionof{table}{Stage-B fitting sensitivity (\%) at $B=512$. The rows share
one Stage-A cache, backbone, codebook size, and evaluation sample sets while
varying the indicated fitting component.}
\label{tab:appendix_training_ablation}
\end{center}

Table~\ref{tab:appendix_training_ablation} is read within its sweep: the first
row anchors the Stage-B variants, while the main table reports the selected
end-to-end operating point, so cross-table subtraction is not a component
effect. The variants remain within 1.20 points on each benchmark, indicating
that ONCE does not require a uniquely tuned learner. We retain source-count
weighting, long-tail correction, and balanced fitting because they preserve
summarized token mass, limit domination by the largest source categories, and
encourage broad codeword use. These choices shape the corpus prior; reusable
grouping and input-specific source-mean aggregation remain the common
mechanism.

\paragraph{Additional ablation results.}
Nearest-source output gives 54.03\% on MVBench and 53.78\% on
LongVideoBench, compared with 54.50\% and 52.95\% for the Full
reference. The changes have opposite signs and provide no consistent
benchmark-wide preference, so we treat nearest-source output as a
representation-sensitivity control. Its mean cosine similarity to the pooled source
representation is 0.944 and 0.924, compared with 0.864 and 0.887 for the
corresponding global codeword prototypes. These diagnostics are accumulated
over 2,047,100 occupied MVBench groups and 684,187 occupied LongVideoBench
groups.

\begin{center}
\centering
\small
\setlength{\tabcolsep}{2.8pt}
\begin{tabular}{@{}llrr@{}}
\toprule
Variant & Diagnostic & MV & LVB \\
\midrule
Full & Coverage (\%) & 100.0 & 100.0 \\
w/o complete reassignment & Coverage (\%) & \AblNoReassignCoverageMV
  & \AblNoReassignCoverageLVB \\
Source-mean output & Cosine to mean & 1.000 & 1.000 \\
Medoid output & Cosine to mean & \AblMedoidFidelityMV & \AblMedoidFidelityLVB \\
Codeword output & Cosine to mean & \AblCodewordFidelityMV
  & \AblCodewordFidelityLVB \\
\bottomrule
\end{tabular}
\captionof{table}{Mechanism diagnostics for the LLaVA-OneVision-7B ablations.
Coverage is the percentage of source tokens represented by output groups;
cosine is similarity to the Full pooled source representation.}
\label{tab:appendix_mechanism_diagnostics}
\end{center}

Medoid outputs stay closer to pooled source means than global codeword
prototypes on both benchmarks, and their accuracies remain near Full. This
joint fidelity and accuracy pattern supports the mechanism used in the main
paper: codewords define group membership, whereas current-video source features
provide the compressed content. The coverage rows answer a separate question.
Complete reassignment guarantees source-token participation, but the mixed
accuracy response shows that coverage alone is not a proxy for task quality.

\paragraph{Inference efficiency.}
Tables~\ref{tab:efficiency} and~\ref{tab:efficiency_05b} report the complete
measurements for the 7B and 0.5B LLaVA-OneVision models, respectively.

\begin{figure*}[!t]
\centering
\includegraphics[width=0.98\textwidth]{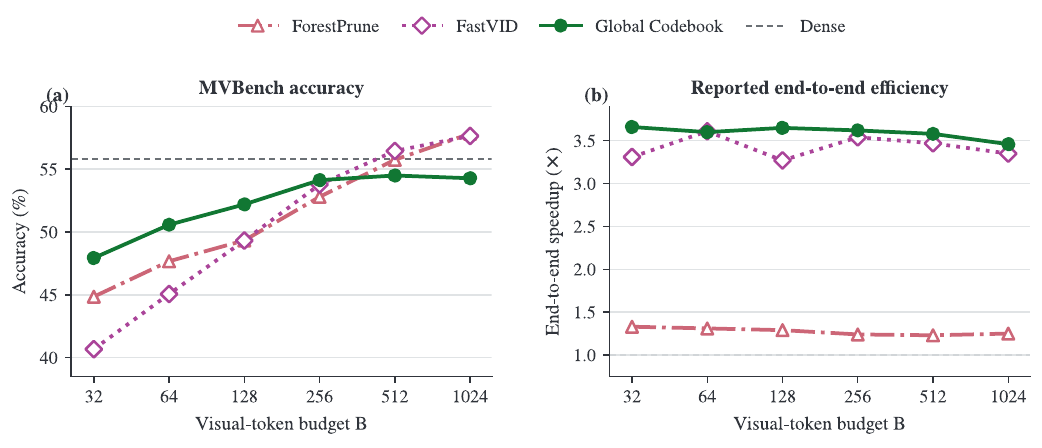}
\caption{Accuracy and reported inference efficiency on
LLaVA-OneVision-7B across visual-token budgets. Panel (a) reports MVBench
accuracy; the dashed line marks the uncompressed result of 55.78\%. Panel (b)
reports end-to-end speedup relative to the Dense latency of 3653.79 ms; the
dashed line marks 1$\times$. All speedups follow the inference setup in
Table~\ref{tab:efficiency}.}
\label{fig:ergc_ablation}
\end{figure*}

\begin{center}
\small
\setlength{\tabcolsep}{3.5pt}
\renewcommand{\arraystretch}{1.05}

\begin{tabular}{@{}lrrrr@{}}
\toprule
Method & \# Tokens & Lat. (ms) & E2E & Prefill \\
\midrule

Dense
& 23,329
& 3653.79
& 1.00$\times$
& 1.00$\times$ \\

\midrule

\multirow{6}{*}{ForestPrune}
& 32
& 2747.38
& 1.33$\times$
& 4.31$\times$ \\
& 64
& 2791.53
& 1.31$\times$
& 4.16$\times$ \\
& 128
& 2836.98
& 1.29$\times$
& 3.72$\times$ \\
& 256
& 2937.61
& 1.24$\times$
& 3.39$\times$ \\
& 512
& 2962.96
& 1.23$\times$
& 2.78$\times$ \\
& 1024
& 2915.80
& 1.25$\times$
& 2.25$\times$ \\

\midrule

\multirow{6}{*}{FastVID}
& 32
& 1102.72
& 3.31$\times$
& 69.58$\times$ \\
& 64
& 1012.68
& 3.61$\times$
& 65.18$\times$ \\
& 128
& 1117.70
& 3.27$\times$
& 52.15$\times$ \\
& 256
& 1032.48
& 3.54$\times$
& 45.59$\times$ \\
& 512
& 1051.65
& 3.47$\times$
& 33.57$\times$ \\
& 1024
& 1092.14
& 3.35$\times$
& 22.73$\times$ \\

\midrule

\multirow{6}{*}{Ours}
& 32
& 998.91
& 3.66$\times$
& 69.34$\times$ \\
& 64
& 1014.50
& 3.60$\times$
& 64.26$\times$ \\
& 128
& 1001.12
& 3.65$\times$
& 60.71$\times$ \\
& 256
& 1010.02
& 3.62$\times$
& 50.15$\times$ \\
& 512
& 1019.36
& 3.58$\times$
& 36.46$\times$ \\
& 1024
& 1056.44
& 3.46$\times$
& 23.92$\times$ \\

\bottomrule
\end{tabular}

\captionof{table}{
Inference efficiency under different visual token budgets on LLaVA-OneVision-7B.
$\#$ Tokens denotes the number of visual tokens.
Lat. is measured in milliseconds, while E2E and Prefill report speedups relative to Dense.
}
\label{tab:efficiency}
\end{center}

Prefill speedup decreases as more pooled tokens are retained, whereas
end-to-end speedup varies over a much narrower range. This gap supports the
runtime interpretation in the main paper: compression removes language-model
prefill work, but visual encoding, compression overhead, and autoregressive
decoding limit the realized wall-clock gain. ONCE preserves this end-to-end
advantage over Dense across the tested budgets, while all ONCE accuracy rows
use the same compression pipeline.

\paragraph{Qwen3.5-9B ablations.}
Table~\ref{tab:qwen_ablation} reports component and codebook-size controls on
all 4,000 MVBench examples at \(B=512\). Each row changes the named choice from
the Full \(K=8192\) configuration.

\begin{center}
\centering
\small
\setlength{\tabcolsep}{4.2pt}
\begin{tabular}{@{}lrr@{}}
\toprule
Variant & Accuracy (\%) & \(\Delta\) \\
\midrule
Full (\(K=8192\))             & 63.25 &  0.00 \\
w/o complete reassignment     & 64.63 & \(+1.38\) \\
Codeword output               & 41.33 & \(-21.93\) \\
Uniform fitting weights       & 61.55 & \(-1.70\) \\
\midrule
\(K=2048\)                    & 63.53 & \(+0.28\) \\
\(K=4096\)                    & 63.60 & \(+0.35\) \\
\bottomrule
\end{tabular}
\captionof{table}{Qwen3.5-9B ablations on MVBench at \(B=512\). All rows use 2-FPS
sampling with at most 256 frames and 4,000 examples. Full uses \(Q=4\),
\(K=8192\), source-count weighting, complete reassignment, and source-mean
output. \(\Delta\) is the accuracy change relative to Full.}
\label{tab:qwen_ablation}
\end{center}

Replacing source means with codeword prototypes lowers accuracy by 21.93
points. The same direction as the LLaVA output control supports a shared design
interpretation across backbones: the learned codebook organizes groups, while
source means retain input-specific content. Uniform fitting weights have a
smaller effect, and changing \(K\) from 8192 to 2048 or 4096 changes accuracy
by at most 0.35 points. Codebook capacity therefore need not translate
linearly into downstream quality at a fixed output budget. Removing complete
reassignment raises accuracy by 1.38 points in this control, again separating
its coverage guarantee from benchmark accuracy.

\paragraph{Budget trade-off.}
Figure~\ref{fig:ergc_ablation} aligns task accuracy and reported end-to-end
efficiency for the same methods and token budgets on LLaVA-OneVision-7B. The
global codebook gives the highest MVBench accuracy at $B=32$, 64, 128, and 256,
while the input-adaptive baselines lead at $B=512$ and 1024. The global
codebook has the highest reported end-to-end speedup at five of the six
budgets, with FastVID leading at $B=64$.
At \(B=64\), FastVID and ONCE differ by only 1.82 ms, or 0.18\% of the
approximately 1.01-s latency, which we treat as practical parity. ONCE is
lower at the other five budgets, so the sweep supports a consistently
low-latency regime without relying on the ordering of one near-tied point.
\FloatBarrier

\paragraph{Smaller model.}
The LLaVA-OneVision-0.5B comparison tests whether the same compression behavior
holds across model scales (Table~\ref{tab:main_05b}).

\begin{center}
\centering
\small
\setlength{\tabcolsep}{2.2pt}
\renewcommand{\arraystretch}{1.08}
\begin{tabular}{@{}clrrr@{}}
\toprule
Budget & Method & MV & V-MME & LVB \\
\midrule
23329 & LLaVA-OV-0.5B & 43.08 & 42.59 & 43.01 \\
\midrule
\multirow{2}{*}{32}
 & VQToken & \textbf{38.10} & \textbf{37.93} & 38.82 \\
 & Global Codebook (Ours) & 37.90 & 35.74 & \textbf{40.84} \\
\midrule
\multirow{2}{*}{64}
 & VQToken & 38.25 & \textbf{37.48} & 38.67 \\
 & Global Codebook (Ours) & \textbf{39.50} & 36.81 & \textbf{42.71} \\
\bottomrule
\end{tabular}
\captionof{table}{Accuracy (\%) under different visual-token budgets on
LLaVA-OneVision-0.5B. MV, V-MME, and LVB denote MVBench, Video-MME, and
LongVideoBench. Bold marks the best compressed result within each budget.}
\label{tab:main_05b}
\end{center}

At \(B=32\), the two compressed methods divide the benchmark leads, whereas
ONCE leads on MVBench and LongVideoBench at \(B=64\). Its LongVideoBench
advantage at both budgets is consistent with global grouping retaining
evidence distributed across longer videos, but the Video-MME rows rule out
uniform dominance. Together with the 7B results, the table supports
applicability across model scales while keeping the accuracy conclusion
benchmark dependent.

% \begin{center}
% \small
% \setlength{\tabcolsep}{1.6pt}
% \renewcommand{\arraystretch}{1.05}
% \begin{tabular}{@{}lrrrrr@{}}
% \toprule
% Method & Tokens & Lat. (ms) & E2E & GiB & Prefill \\
% \midrule
% Dense & 23329 & 698.50 & -- & 22.13 & -- \\
% VQToken & 32 & 555.60 & 1.26$\times$ & 5.74 & 10.19$\times$ \\
% VQToken & 64 & 601.70 & 1.16$\times$ & 5.74 & 10.13$\times$ \\
% Ours & 32 & 529.00 & 1.32$\times$ & 5.53 & 9.92$\times$ \\
% Ours & 64 & 511.60 & 1.37$\times$ & 5.53 & 10.13$\times$ \\
% \bottomrule
% \end{tabular}
% \captionof{table}{Inference efficiency on LLaVA-OneVision-0.5B. Speedups are
% relative to Dense; -- denotes the reference value.}
% \label{tab:efficiency_05b}
% \end{center}
\begin{center}
\small
\setlength{\tabcolsep}{3.5pt}
\renewcommand{\arraystretch}{1.05}

\begin{tabular}{@{}lrrrr@{}}
\toprule
Method & \# Tokens & Lat. (ms) & E2E & Prefill \\
\midrule

Dense
& 23,329
& 1362.54
& 1.00$\times$
& 1.00$\times$ \\

\midrule

\multirow{2}{*}{VQToken}
& 32
& 1108.76
& 1.23$\times$
& 14.40$\times$ \\
& 64
& 1263.03
& 1.08$\times$
& 14.10$\times$ \\

\midrule

\multirow{2}{*}{Ours}
& 32
& 985.26
& 1.38$\times$
& 14.00$\times$ \\
& 64
& 984.72
& 1.38$\times$
& 14.14$\times$ \\

\bottomrule
\end{tabular}

\captionof{table}{Inference efficiency on a uniform 400-example MVBench subset
with LLaVA-OneVision-0.5B. E2E and Prefill are speedups over Dense.}
\label{tab:efficiency_05b}
\end{center}

ONCE has the higher end-to-end speedup at both budgets even though the two
methods have similar prefill speedups. This separation points to costs outside
language-model prefill and reinforces the main-paper distinction between
token-level computation and measured latency. ONCE's nearly unchanged latency
between \(B=32\) and \(B=64\) further suggests that fixed inference stages
dominate this small budget change on the 0.5B backbone.

\begin{center}
\centering
\small
\setlength{\tabcolsep}{2.8pt}
\begin{tabular}{@{}llrrr@{}}
\toprule
& & \multicolumn{2}{c}{Peak reserved (GiB)} & Reduction \\
\cmidrule(lr){3-4}
Model & Data & Dense & ONCE & (\%) \\
\midrule
LLaVA-OV-7B & MV  & 23.89 & 19.99 & 16.36 \\
LLaVA-OV-7B & LVB & 24.01 & 20.00 & 16.69 \\
Qwen3.5-9B  & MV  & 37.73 & 31.85 & 15.58 \\
Qwen3.5-9B  & LVB & 41.42 & 35.46 & 14.39 \\
\bottomrule
\end{tabular}
\captionof{table}{Absolute peak GPU memory for the matched Dense and ONCE runs at
$B=512$. MV and LVB denote MVBench and LongVideoBench. Values are PyTorch peak
reserved memory; all runs use batch size 1 on the same local 48-GiB RTX~4090.}
\label{tab:absolute_peak_memory}
\end{center}

ONCE lowers peak reserved memory in all four matched pairs, despite the
different absolute footprints of the two backbones and benchmarks. The
14.39--16.69\% memory reductions are much smaller than the corresponding
visual-token reductions in the main paper, which is consistent with model
weights and non-visual state remaining resident. The table therefore supports
memory savings as a practical consequence of compression without equating
token reduction with total device-memory reduction.

% Check whether the conference requires a reproducibility checklist to be included in the paper.
% If so, you can uncomment the following line and ajust the path to include it.
% \input{ReproducibilityChecklist.tex}

\end{document}